\documentclass[sigconf]{acmart}

\copyrightyear{2022}
\acmYear{2022}
\setcopyright{acmcopyright}
\acmConference[KDD '22] {Proceedings of the 28th ACM SIGKDD Conference on Knowledge Discovery and Data Mining}{August 14--18, 2022}{Washington, DC, USA.}
\acmBooktitle{Proceedings of the 28th ACM SIGKDD Conference on Knowledge Discovery and Data Mining (KDD '22), August 14--18, 2022, Washington, DC, USA}
\acmPrice{15.00}
\acmISBN{978-1-4503-9385-0/22/08}
\acmDOI{10.1145/3534678.3539395}

\usepackage{amsmath, amsfonts, amsthm, xspace, color}
\usepackage{booktabs} 

\usepackage{multirow, tabularx} 
\usepackage{booktabs} 
\usepackage{enumitem}
\usepackage[T1]{fontenc}
\usepackage{epstopdf}
\usepackage{graphicx}
\usepackage{url}
\usepackage{wrapfig,lipsum}
\usepackage{makecell}
\usepackage{wrapfig}
\usepackage{latexsym}

\usepackage{microtype}

\usepackage{caption}
\usepackage{subcaption}

\newcommand{\red}[1]{\textcolor{red}{#1}}

\definecolor{darkgreen}{rgb}{0.0, 0.5, 0.0}
\newcommand{\green}[1]{\textcolor{darkgreen}{#1}}
\newcommand{\hide}[1]{} 

\newcommand{\ie}{i.e.\xspace} 
\newcommand{\eg}{e.g.\xspace} 
\newcommand{\nop}[1]{}

\newcommand{\naive}{na\"{\i}ve\xspace} 

\newtheoremstyle{exampstyle}
  {0pt} 
  {0pt} 
  {\itshape} 
  {1em} 
  {\bfseries} 
  {.} 
  {.5em} 
  {} 

\theoremstyle{exampstyle}
\newtheorem{thm:def}{Definition}

\newtheorem{thm:eg}{Example}
\newtheorem{thm:lem}{Lemma}
\newtheorem{thm:obs}{Observation}
\newtheorem{thm:req}{Requirement}
\newtheorem{thm:prop}{Proposition}
\newtheorem{thm:principle}{Principle}
\newtheorem{thm:thm}{Theorem}
\newtheorem{thm:corollary}{Corollary}

\newcommand{\abs}[1]{\mathopen| #1 \mathclose|}			

\def \C {\mathcal{C}}
\def \D {\mathcal{D}}

\def \G {\mathcal{G}}

\def \K {\mathcal{K}}

\def \T {\mathcal{T}}

\newcommand{\Our}{\mbox{EvMine}\xspace}
\newcommand{\OurNoClass}{\mbox{EvMine-NoClass}\xspace}
\newcommand{\OurCOOC}{\mbox{EvMine-COOC}\xspace}
\newcommand{\OurNoLM}{\mbox{EvMine-NoLM}\xspace}
\newcommand{\OurSingle}{\mbox{EvMine-Single}\xspace}

\usepackage{adjustbox}
\usepackage[super]{nth}
\usepackage{dsfont}

\begin{document}

\title{Unsupervised Key Event Detection from Massive Text Corpora}



\settopmatter{authorsperrow=4}

\author{Yunyi Zhang}
\affiliation{%
  \institution{UIUC, IL, USA}
  \country{}}
\email{yzhan238@illinois.edu}

\author{Fang Guo}
\affiliation{%
  \institution{Westlake University, China}
  \country{}}
\email{guofang@westlake.edu.cn}

\author{Jiaming Shen}
\affiliation{%
  \institution{Google Research, NY, USA}
  \country{}}
\email{jmshen@google.com}

\author{Jiawei Han}
\affiliation{%
  \institution{UIUC, IL, USA}
  \country{}}
\email{hanj@illinois.edu}

\renewcommand{\shortauthors}{Yunyi Zhang et al.}


\begin{abstract}

Automated event detection from news corpora is a crucial task towards mining fast-evolving structured knowledge.
As real-world events have different granularities, from the top-level themes to key events and then to event mentions corresponding to concrete actions, there are generally two lines of research:
(1) \emph{theme detection} tries to identify from a news corpus major themes (\eg, ``2019 Hong Kong Protests'' versus ``2020 U.S. Presidential Election'') which have very distinct semantics; and
(2) \emph{action extraction} aims to extract from a single document mention-level actions (\eg, ``the police hit the left arm of the protester'') that are often too fine-grained for comprehending the real-world event.
In this paper, we propose a new task, \emph{key event detection} at the intermediate level, which aims to detect from a news corpus key events (\eg, \textsc{HK Airport Protest on Aug. 12-14}), each happening at a particular time/location and focusing on the same topic.
This task can bridge event understanding and structuring and is inherently challenging because of (1) the thematic and temporal closeness of different key events and (2) the scarcity of labeled data due to the fast-evolving nature of news articles.
To address these challenges, we develop an unsupervised key event detection framework, \Our, that
(1) extracts temporally frequent peak phrases using a novel ttf-itf score,
(2) merges peak phrases into event-indicative feature sets by detecting communities from our designed peak phrase graph that captures document co-occurrences, semantic similarities, and temporal closeness signals, and
(3) iteratively retrieves documents related to each key event by training a classifier with automatically generated pseudo labels from the event-indicative feature sets and refining the detected key events using the retrieved documents in each iteration.
Extensive experiments and case studies show \Our outperforms all the baseline methods and its ablations on two real-world news corpora.

\end{abstract}



\maketitle

\section{Introduction}
\begin{figure}[t]
    \centering
    \centerline{\includegraphics[width=0.45\textwidth]{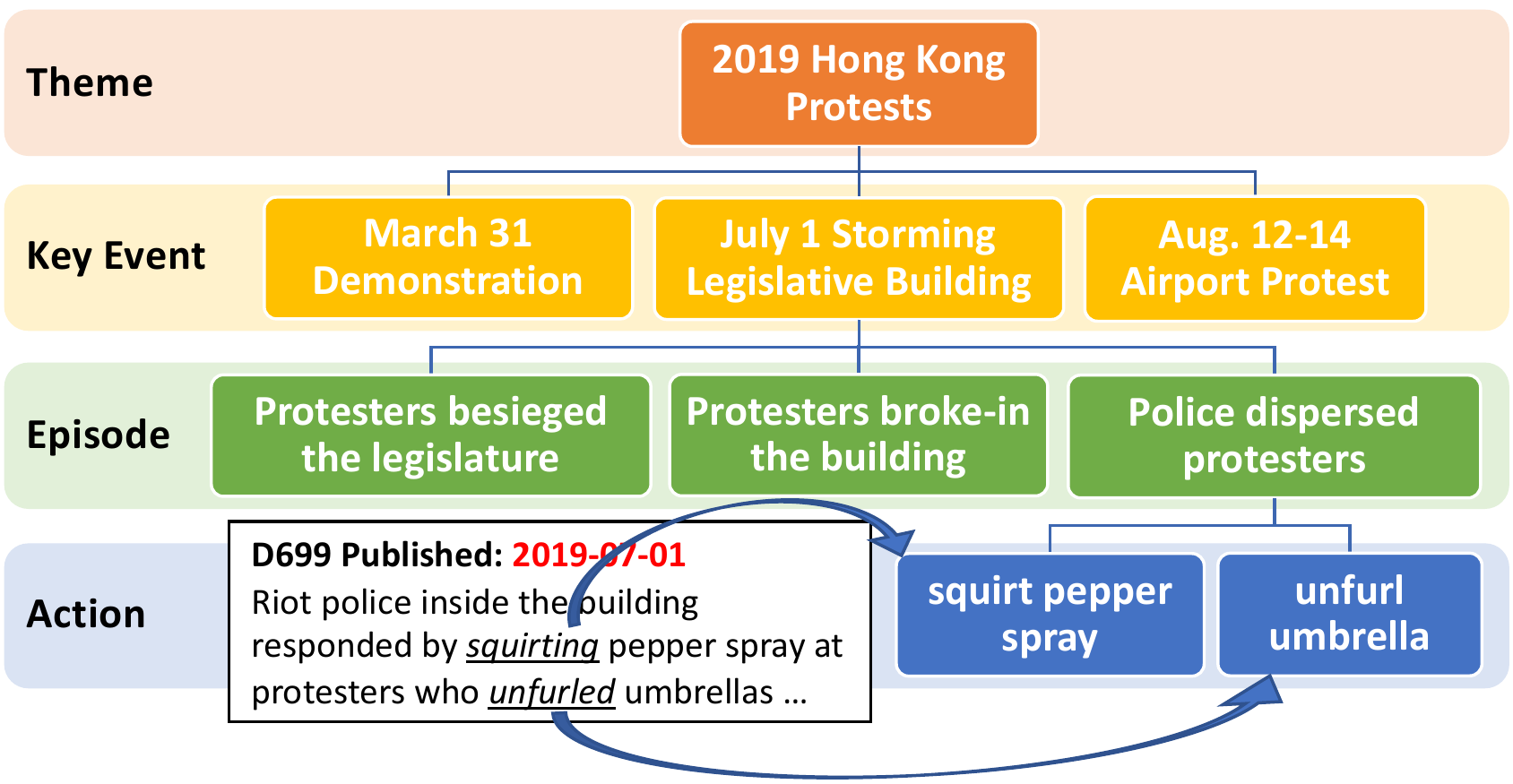}}
    \vspace*{-1.5em}
    \caption{An example of event structure hierarchy.}
    \label{fig:event_taxo}
    \vspace*{-2em}
\end{figure}

Automated real-world event discovery has long been studied to help people quickly digest explosive information.
Researchers studying human memory find people tend to organize real-world events in a hierarchical way~\cite{wheeler1997toward, Burt2003ThemesEA}, ranging from top-level \emph{themes} (\eg, ``2019 Hong Kong (HK) Protests'') to middle level \emph{key events} (\eg, \textsc{July 1 Storming Legislative Building}), possibly to sub-middle level \emph{episodes} (\eg, ``Protester besieged the legislature''), and down to bottom level \emph{actions}\footnote{\small We define actions as text spans, and multiple actions may refer to the same real-world action, which is related to the event coreference resolution task. Actions and key events are also called event mentions and super-events in many NLP literature. As many studies do not distinguish these two levels of abstraction, we propose a clear event structure for systematic analysis.} 
(\eg, ``Riot police squirt pepper spray at protesters'').
As shown in Figure~\ref{fig:event_taxo}, going up this event structure hierarchy leads to larger and more coarse-grained ``events'' whereas moving down the hierarchy brings in more fine-grained and concrete mentions of ``events''.

\begin{figure}[!t]
    \centering
    \centerline{\includegraphics[width=0.42\textwidth]{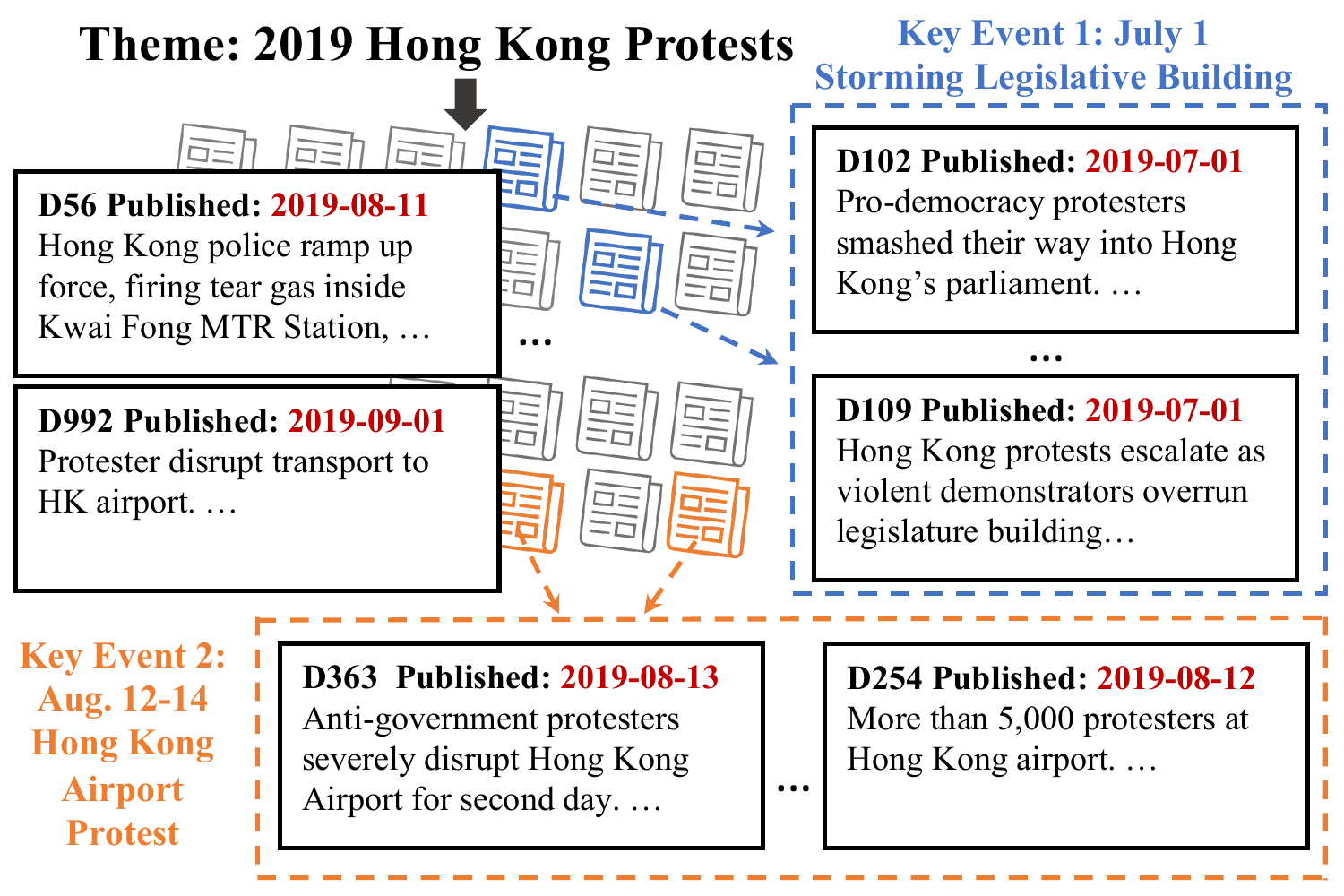}}
    \vspace*{-1em}
    \caption{An illustrative example of key event detection task.}
    \label{fig:intro}
    \vspace*{-1.5em}
\end{figure}

The broad spectrum of ``events", differing in duration and complexity, has fostered a variety of event discovery studies under different task names. 
One line of research, named Topic Detection and Tracking (TDT)~\cite{Allan1998TopicDA,Becker2010LearningSM,Xu2019ResearchOT}, aims to detect themes from an input corpus where each theme is represented by a cluster of documents, focusing on distinct thematic topics.
For example, documents about ``2019 Hong Kong Protests'' are thematically very distinct from those of ``2020 U.S. Presidential Election''.
As a result, content-based document clustering methods~\cite{Allan1998OnlineNE, Sayyadi2013AGA} can easily separate those themes.
However, these methods cannot effectively distinguish the key events of the same/similar themes (\eg, identify documents about \textsc{HK Legislative Building Storming} and \textsc{HK Airport Sit-In} from a corpus related to ``2019 HK protests'')~\cite{ge-etal-2016-event}.
Another line of work~\cite{Chen2015EventEV,Nguyen2018GraphCN,Du2020EventEB,Shen2021ETypeClus} is action extraction which tries to extract concrete actions (represented as text spans) from input documents.
For example, one action extracted from the sentence in Figure~\ref{fig:event_taxo} can be ``Riot police squirt pepper spray at protesters''.
These methods typically require a predefined event schema along with massive human-labeled documents for model learning.
Besides, their output event mentions are highly redundant as one real-world event can usually be expressed in different ways in multiple documents, which further prevents humans from seeing the overall picture of the event.

In this paper, we propose a new task, \emph{key event detection}, which aims to detect key events from a corpus about one general event theme, or \emph{theme corpus}.
We assume each document has a publication date in the corpus, and each key event, as an aggregation of actions with concrete event time and focused location, is usually covered by a collection of documents.
Since previous studies work on either too coarse or too fine-grained views of events, key event discovery, sitting in the intermediate level, plays an essential role in bridging the real-world event understanding and structuring.
As shown in Figure~\ref{fig:intro}, given a theme corpus about ``2019 Hong Kong Protests'', we extract key events such as \textsc{July 1st Storming Legislative Building} and \textsc{Aug. 12-14 Hong Kong International Airport Protest}\footnote{\small All key events in this paper are manually named for easy understanding.},
which helps people gain insights about the theme and thus compensates the previous TDT studies.
Meanwhile, first detecting key events provides subsequent action extraction models with extra clues on what type of actions will most likely appear in each document. 
Furthermore, the identified key events can directly benefit many downstream tasks like timeline generation~\cite{Ge2015BringYT}, evolutionary analysis~\cite{Yang2006TracingTE}, and query expansion~\cite{Rosin2021EventDrivenQE}.

Our proposed key event discovery task, while being useful, has its own challenges. 
First, compared to previous topic detection and tracking task, our task is intrinsically harder because it aims to distinguish key events of the same theme and those key events are often thematically similar and temporally closer to each other.
Besides, as new events are happening every day, it is neither realistic nor scalable to curate all event schema in advance or label documents for training supervised event extraction models.

To address the above challenges, we propose an unsupervised key event detection framework, \Our, that requires no human-labeled training data and can automatically discover key events from a large theme corpus.
\Our contains three major steps.
First, we extract event-related ``peak phrases'' from input corpus based on our proposed ``temporal term frequency--inverse time frequency'' (ttf-itf) measure,
where each peak phrase is unusually frequent on a day and thus likely indicates a key event.
Second, as some key events can span multiple consecutive days and people have various ways to communicate the same key event, we group detected peak phrases into semantic clusters, which will serve as event-indicative features for selecting key event documents.
Specifically, we propose a novel topic-time integrated peak phrase graph that considers document co-occurrence features, pre-trained masked language model (MLM) based semantic similarities, and temporal closeness, based on which we cluster peak phrases with a community detection algorithm.
Each peak phrase cluster corresponds to one key event and provides event-indicative features.
Finally, for each key event, we use the phrases in its corresponding peak phrase cluster to train a classifier that predicts whether a document is related to this key event. 
We also introduce a feedback loop that uses the current classification results to improve the phrase-based pseudo labels and find possibly missing key events, leading to an iterative document selection process with automatic refinement in each iteration.

To summarize, our contributions are:
(1) We introduce a new research problem \emph{key event detection}, which takes a set of documents related to the same theme as inputs and outputs multiple important key events along with their associated documents.
(2) We propose \Our, a novel unsupervised framework for key event detection. \Our automatically extracts temporally frequent peak phrases, clusters them with a graph-based method that combines thematic and temporal information, and applies document classification with iterative refinements to retrieve the most relevant documents for each key event.
(3) We conduct quantitative evaluation, case studies, and parameter sensitivity analysis on two real-world event theme corpora, and \Our outperforms all the baseline methods and its own ablations in terms of the ability to detect key events.
 
\section{Preliminaries} \label{sec:prelim}

\subsection{Problem Definition}

\begin{thm:def}[Key Event] \label{def:key}

    A key event is a set of thematically coherent documents about a particular theme and corresponds to a real-world event that happened at a particular time and/or location.

\end{thm:def}

This work focuses on extracting key events from a news corpus about one theme, which helps people understand the overall plot of this theme and extract useful knowledge from this corpus.
We observe that different people use different ways to describe a key event, but their key phrases are often the same or synonymous and represent the most important aspects of a key event, \eg, times, locations, and topics. 
Such event-indicative phrases will form peaks of frequency in the temporal dimension that can identify key events, which motivates the following definition.

\begin{thm:def}[Peak Phrase] \label{def:peak}

    A peak phrase $(p, t)$ is a tuple of a phrase $p$ and a date $t$, representing
    $p$ is unusually frequent on $t$
    and thus likely comes from a key event happening on that day.

\end{thm:def}
We use phrases, instead of single-token words, as features because they often carry richer semantic information.
We discuss how to extract those phrases in Section~\ref{sec:prepro}.

\begin{thm:def}[Key Event Detection] \label{def:prob_def}

    Given a corpus $(\D, \T)$ about one theme, where each document $d \in \D$ is a news article with its publication date $t(d) \in \T$, our goal is to obtain a set of document clusters $\K = \{K_1, K_2, \dots, K_M\}$ where each cluster $K_i \subset \D$ represents a key event and every two clusters are non-overlapping (i.e., $K_i \cap K_j = \emptyset$ for $i \neq j$).

\end{thm:def}

Note that we may not know the number of key events $M$ in advance and do not require the union of all documents to be equal to the input corpus. Also, although there can be some documents discussing more than one key event, our goal is to mine those distinctive documents for each detected key event and thus require the key events to be non-overlapping.

\vspace{-0.75em}
\subsection{Theme Corpus Retrieval} \label{sec:retrieval}

Although out of scope of our work, we briefly describe how we collect the input theme corpus from massive news data for completeness.
Given an initial massive news corpus, we utilize classic information retrieval ranking function BM25 with heuristic pruning rules. 
For example, for the corpus \textbf{HK Protest}, we use keywords ``Hong Kong'' and ``protest'' as query and prune the retrieved documents if they are not published from Mar. to Nov. 2019. 
Although several previous studies on TDT task are built for similar goal of retrieving corpus under the same theme, we still apply our straightforward strategy since empirically it can achieve a high recall that is sufficient for our task.
Worth noting that, although this simple ranking method may introduce some noisy documents not belonging to any key event, the goal of our task is to extract the theme-related key events only, which differentiates it from simply clustering documents in a clean theme corpus.

\section{Methodology} \label{sec:method}

\begin{figure*}[!t]
    \centering
    \centerline{\includegraphics[width=0.98\textwidth]{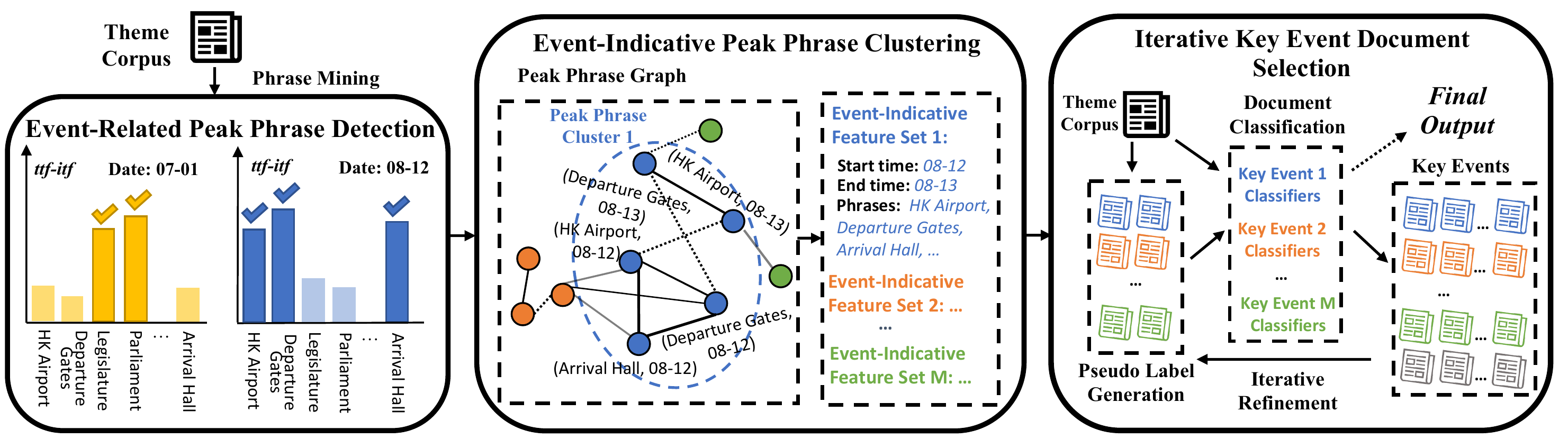}}
    \vspace*{-1em}
    \caption{An overview of our proposed unsupervised key event detection framework \Our. }
    \label{fig:framework}
    \vspace*{-1.5em}
\end{figure*}

We propose a novel unsupervised framework \Our to address the key event detection task. 
As shown in Figure~\ref{fig:framework}, \Our consists of three major components: (1) \emph{an event-related peak phrase detection module} which extracts a set of peak phrases from the input corpus, (2) \emph{an event-indicative peak phrase clustering module} which uses a novel topic-time integrated graph structure to merge the extracted peak phrases into several event-indicative feature sets, and (3) \emph{an iterative key event document selection module} which iteratively trains classifiers using automatically generated pseudo-labels to retrieve event documents and refines the detected key event with the current results.
Below we first describe how we preprocess the input corpus to mine quality phrases and then present each \Our component.

\vspace{-0.75em}
\subsection{Quality Phrase Mining} \label{sec:prepro}

We first extract phrases from the theme corpus to serve as features for key event detection. 
In this work, we apply an unsupervised phrase mining tool, UCPhrase~\cite{gu2021ucphrase}, to extract quality phrases.
UCPhrase first generates ``silver labeled'' phrases by applying frequent pattern mining on the corpus and then trains a lightweight CNN model on the attention weight matrices generated by a pre-trained language model. Since we work on text corpora about an overall theme, we train the UCPhrase model on each corpus from scratch so that it can mine more corpus-specific phrases.

As UCPhrase only considers the quality of phrases, the extracted phrases may not be equally important for identifying key events. Specifically, some phrases may be too general and thus likely represent the background theme (\eg, ``Hong Kong'' for ``2019 Hong Kong Protests''), while some other phrases may be too rare and thus hardly come from any key event. Therefore, we apply a corpus-level tf-idf score to filter out those uninformative phrases:
\vspace{-0.25em}
\begin{equation}
    \small
    \text{tf-idf}(p) = \left(1 + \log{\sum_{d \in \D} freq_d (p)}\right) \log \left( \frac{\abs{\D}}{\abs{\{d \in \D| freq_d(p) > 0\}}} \right)
    \vspace{-0.25em}
\end{equation}
where $freq_d(p)$ represents the frequency of a phrase $p$ in the document $d$ and $\D$ is all the documents in the corpus. Finally, we choose those phrases with the highest $70\%$ corpus-level tf-idf scores as our candidate phrases, from which we will select peak phrases for key event detection. Our framework is not sensitive to this threshold around this value ($60\%\sim80\%$) since we will further select temporally frequent peak phrases as key event-indicative features. 

\vspace{-0.5em}
\subsection{Event-Related Peak Phrase Detection} \label{sec:peak}

We observe that key events are often discussed by multiple news articles and thus those key event indicative phrases appear more frequently on some specific days compared to the entire corpus.
For example, the phrase ``Hong Kong Airport'' occurs in most of the documents about \textsc{Aug. 12-14 Hong Kong Airport Protest}, and thus it can help us detect documents about this key event.

Based on this observation, we propose to first detect these key event related phrases, defined as ``peak phrases'', to serve as features for key events. As stated in Definition~\ref{def:peak}, a peak phrase consists of a phrase and a date when the phrase is unusually frequent. This means that a peak phrase $(p, t)$ should meet the following two criteria: (1) $p$ should be frequent on $t$ and (2) $p$ should not be equally common for the entire corpus but only frequent on some specific days. In order to select such peak phrases, we modify the traditional tf-idf and define a novel ``temporal term frequency--inverse time frequency'' (ttf-itf) measure to rank the candidate phrases. 

\smallskip
\noindent \textbf{Temporal term frequency.}
Temporal term frequency (ttf) measures how frequent a phrase is on a day. A \naive way is to directly use the raw frequency of this phrase on this day. However, we observe that there are always delays and back referencing in news articles, meaning that for those key events, news articles will continue discussing them for several days (or even months) after they happened. Therefore, we propose to aggregate frequencies from several later days when calculating temporal term frequency. Specifically, we define ttf of a phrase $p$ on a day $t$ as follows:
\vspace{-0.25em}
\begin{equation}
    \small
    \text{ttf}(p, t) = \frac{1}{n_t} \sum_{i=0}^{n_t-1} \left(1 - \frac{i}{n_t}\right) freq_{t+i}(p),
\vspace{-0.25em}
\end{equation}
where $freq_t(p)$ is the raw frequency of phrase $p$ on the day $t$, and $n_t$ is a parameter representing how many following days we would like to aggregate term frequencies from. 
This formulation helps selecting phrases discussed continuously for a period and thus likely identifies a key event.
For example, ``Hong Kong Airport'' will have a high ttf score on some specific days, since it is continuously discussed during and after the key event \textsc{Aug. 12-14 Airport Protest}. Also, we use decreasing weights when aggregating frequencies to favor those days with high frequency by themselves, because these are the days when a key event likely happens. 

\smallskip
\noindent \textbf{Inverse time frequency.}
Besides the requirement of being frequent on someday, a peak phrase should also be uncommon for the entire corpus. To measure how uncommon a phrase is on the day level, we propose inverse time frequency (itf) as follows:
\begin{equation}
    \small
    \text{itf}(p) = \frac{\max \T - \min \T + 1}{\abs{\{t \in \T | freq_t(p) > 0\}}},
\end{equation}
where the numerator is the total number of days covered by the corpus, including those days without any document, and the denominator is the number of days when phrase $p$ is mentioned. 
Intuitively, if a phrase is indicative of a key event, then it will only be mentioned frequently around the event happening time, leading to a high itf score. However, during the event happening time, this phrase will occur in multiple documents and possibly have low idf score, meaning that classic idf score is not proper for detecting key event indicative phrases.

Finally, the ttf-itf of $(p, t)$ combines the above two measures: 
\vspace{-0.25em}
\begin{equation}
    \small
    \text{ttf-itf}(p, t) = \text{ttf}(p, t) \cdot \log \text{itf}(p),
\vspace{-0.25em}
\end{equation}
based on which we can generate a ranked list of possible peak phrases and select the most confident ones to further group them into event-indicative peak phrase sets. For example, some top-ranked peak phrases of ``2019 Hong Kong Protests'' theme include ``Hong Kong Aiport'' on Aug 12 and Aug 13 representing the \textsc{Aug. 12-14 Airport Protest} event, and ``legislative council'' on July 1 indicating the \textsc{July 1 Storming Legislative Building} event.

Note that this method can only select those phrases that are frequent on some day, but the wording of news articles can be different, so we may miss some phrases that are possibly indicative of some key event but relatively infrequent. For example, ``Hong Kong Airport'' is interchangeable with ``Hong Kong International Airport'', but one of them may be used more frequently than the other. However, this will not be a problem, since our event-indicative peak phrase clustering step (c.f. Sect.~\ref{sec:cluster}) aims to cluster the selected peak phrases into event-indicative feature sets and the frequent phrases are enough for identifying key events. Also, our iterative key event document selection module (c.f. Sect.~\ref{sec:doc_class}) will train text classifiers to retrieve those key event articles with different wordings.

\vspace{-0.5em}
\subsection{Event-Indicative Peak Phrase Clustering} \label{sec:cluster}

After selecting peak phrases from the corpus, we aim to group them into clusters, each of which serves as features for a candidate key event. Specifically, we need to consider (1) whether two peak phrases on the same day come from the same key event and (2) whether two peak phrases from consecutive days come from the same key event. 
The former helps distinguish key events with overlapping time periods, while the latter is necessary for detecting events lasting for several days. 
In this section, we present a novel graph structure that uses peak phrases as nodes and seamlessly combines phrase co-occurrence features, pre-trained masked language model (MLM) based semantic similarities, and the temporal closeness into the edge weights for better feature generation tailored for the key event detection task.

\smallskip
\noindent \textbf{Pre-trained MLM-based phrase embeddings.} We first describe how we get the MLM-based embedding for each phrase to capture both content and context features simultaneously. Suppose a phrase $p$ appears $N_p$ times in the corpus. Then, for each of its mention $p^l$, $l \in \{1, 2, \dots, N_p\}$, we obtain its \emph{content feature} $\mathbf{x}^l_p$ by feeding the original sentence into a pre-trained MLM and taking the average of the generated embedding vectors corresponding to the tokens of $p$. To get the \emph{context feature} $\mathbf{y}^l_p$ of this mention, we first replace the entire phrase $p$ with a single [\textsc{Mask}] token, feed the new sentence into the same language model, and then use the embedding vector of this [\textsc{Mask}] token as the context feature. Finally, to get the phrase embedding that captures both content and context features, we concatenate two feature vectors for each mention and take the average of the resulting mention vectors:
\vspace{-0.25em}
\begin{equation}
    \small
    \mathbf{e}_p = \frac{1}{N_p} \sum_{l = 1}^{N_p} [\mathbf{x}^l_p; \mathbf{y}^l_p].
\vspace{-0.25em}
\end{equation}

\smallskip
\noindent \textbf{Peak phrase graph construction.}
Each node in the peak phrase graph corresponds to a peak phrase $n_i = (p_i, t_i)$ selected in the previous step. There are two types of edges connecting these nodes:
\begin{enumerate}[wide, labelwidth=0pt, labelindent=0pt]
    
    \item For each pair of same-day peak phrases, we connect them with an edge whose weight is decided by their document co-occurrences and LM-based semantic similarity.
    The document co-occurrence features are captured with the normalized pointwise mutual information, or npmi score (formula in Appx.~\ref{app:npmi}).
    Basically, npmi compares the probability of a document mentioning two phrases with the expected value, so a large npmi score implies that they do not co-occur ``by chance'' and thus more likely correspond to the same key event. For example, on Aug 12, 2019, the phrase ``Hong Kong Airport'' likely has a higher npmi score with ``Arrival Hall'' than ``Victoria Park'', since they come from the same key event \textsc{Aug. 12-14 Hong Kong Airport Protest}.
    
    Besides, the semantic similarity between two phrases is calculated as the cosine similarity between their pre-trained MLM-based embeddings. The semantic similarity complements npmi score since if an entity has multiple synonymous surface names (\eg, ``Hong Kong Airport'' and ``Hong Kong International Airport''), a single news article tends to use only one of them due to clarity or writing habits, leading to small npmi scores between synonymous phrases. Therefore, compensating the npmi scores with embedding-based similarity scores helps boosting the edge weights between synonyms. The final edge weight between two nodes $n_i$ and $n_j$ is the geometric mean of the above two measures after truncating their negative values:
\vspace{-0.25em}
    \begin{equation}
        \small
        w_{i,j} = \sqrt{\max\Big(\text{npmi}(n_i, n_j), 0\Big) \cdot \max\Big(\text{CosSim}(e_{p_i}, e_{p_j}), 0\Big)}
\vspace{-0.25em}
    \end{equation}
    
    \item For two peak phrases on consecutive days, we assume that they likely come from the same key event if they correspond to the same phrase, since for events lasting for several days, their most indicative phrases will also be continuously mentioned by multiple news articles in this period. Notice that the maximum value of the first type of edge weight is 1, so to reflect our assumption, we set the edge weight between such a pair of peak phrases $(p, t)$ and $(p, t+1)$ a constant value $w > 1$. By doing so, when later we detect communities on the graph, the algorithm will be forced to put same-phrase consecutive-day peak phrases into the same cluster so that it can detect multi-day key events. An example could be connecting the phrase ``Hong Kong Airport'' within the period Aug. 12-14 for detecting the event \textsc{Aug. 12-14 Hong Kong Airport Protest}.
\end{enumerate}

\smallskip
\noindent \textbf{Peak phrase clustering.}
After constructing the peak phrase graph, we apply a lightweight community detection algorithm, Louvain~\cite{blondel2008FastUO}, to generate non-overlapping communities in this graph.
At a high level, Louvain detects graph communities by maximizing the density of links within communities compared to links between communities.\footnote{We use Louvain algorithm since it is efficient and does not need the number of clusters. In principle, other kinds of community detection methods or graph embeddings plus clustering algorithms can be applied, which we leave as future works.}
We keep those communities with at least two nodes, and each remaining community, denoted as $\C_i = (P_i, T_i)$, contains a cluster of peak phrases $P_i$ and a time span $T_i$.
Since most news articles will discuss key events right after they happen, we can view the time span $T_i$ as an estimation of the key event happening time.
Figure~\ref{fig:framework} shows an example. 
The detected community in the peak phrase graph includes some event-indicative phrases like ``Hong Kong Airport'', ``arrival hall'', and ``departure gates'' as well as an estimated key event happening time Aug. 12 to Aug. 13, 2019.

\subsection{Iterative Key Event Document Selection} \label{sec:doc_class}
The previous step generates a feature set of event-indicative phrases and an estimated time period for each key event. 
In this step, we aim to further classify documents from the corpus into these events. As discussed in Section~\ref{sec:peak}, since we select the peak phrases based on their frequencies, the detected peak phrase clusters will not include any infrequent key phrases. Therefore, if we directly apply document-phrase matching to retrieve event-related documents, we will likely miss those documents with different wordings. However, phrase matching can still help identify some high-quality event-related documents, so we propose an iterative key event document selection module that generates pseudo-labeled documents using phrase matching, trains text classifiers to retrieve event documents, and then applies a feedback loop to iteratively refine the event-indicative features with the current results.

Before training the classifier, we first get the document embeddings using an LM-based sentence encoder, Sentence Transformer~\cite{reimers-2019-sentence-bert}, which is a transformer model that is first pre-trained on the masked language model task~\cite{Devlin2019BERTPO} and then fine-tuned with a sentence similarity objective. Specifically, we take the lead-3 sentences for each document, feed them individually into a pre-trained Sentence Transformer encoder, and take the average of the resulting 3 sentence embedding vectors as the document representation. Here, we only use the lead-3 sentences, since news articles tend to summarize the event at the very beginning. Such a strategy has been used in previous related work~\cite{Ribeiro2017UnsupervisedEC, Shang2018InvestigatingRN} and also serves as a strong baseline for the text summarization task~\cite{See2017GetTT}.

\smallskip
\noindent \textbf{Pseudo label generation.}
As just stated, for each key event, we first find from the corpus all documents published in the estimated event time. 
We also enrich the event-indicative phrases by adding in obvious synonymous phrases that have embedding similarities greater than a threshold $\tau$ with any of the current event-indicative phrases (\eg, $\tau=0.95$ in our experiments).
Then, for each document, we count the number of times it mentions any of the enriched event-indicative phrases for this key event. 
Finally, we rank these documents by their phrase matching results and select several top-ranked documents as pseudo-labeled documents for this key event, denoted by $K^0_i$. Since we want to select high-quality positive documents, we only use the most confident ones (\eg, top-10 in our experiments) as pseudo-labeled samples, and we will introduce how we train the classifiers with such a limited number of labels.

\smallskip
\noindent \textbf{Classifier training.}
Since our task is class-unbalanced classification with null class, we propose to train individual binary classifiers for each detected key event, instead of applying some multi-class models that are mostly designed for the class-balanced task without null class.
Since there are much more negative documents than positive ones in the corpus, which is similar to the settings of the entity set expansion task, we follow the training strategy proposed in~\cite{Shen2020SynSetExpanAI}. Specifically, we randomly sample $r$ times the number of pseudo-labeled positive documents for each event from the given corpus and train an SVM classifier on these $(r+1)\cdot\abs{K^0_i}$ samples. Then, we repeat the above process by $S$ times for robustness, and we take the average of the prediction scores from the above $S$ classifiers to get the final prediction score for how likely a document is related to a key event. Since each document will have a prediction score for each detected key event, we first assign a document to the key event with the highest score if it is positive. Then, for each key event, we rank its assigned documents by their scores, which gives us a ranked list of the most confident documents for this key event.

\smallskip
\noindent \textbf{Post-classification processing.}
The above learned document classifiers rely only on the text information.
We propose to further leverage the document temporal information to refine the initial classification results in a post-processing step. 
As each key event has an estimated time span $T_i$ (c.f. Section~\ref{sec:cluster}),  one naive approach is to directly filter out those documents that are not published within $T_i$.
However, this approach could be problematic as some news articles published outside the estimated time period can still talk about the target key event via afterward comments or discussing newly exposed information.
To better utilize document time information, we first group all the documents published outside $T_i$ into consecutive-day clusters.
Two documents will reside in the same cluster if they are published on the same day or in consecutive days.
Then, we filter out the document that is published outside $T_i$ if it (1) belongs to a cluster with multiple documents or (2) does not explicitly express a timestamp in its lead-3 sentences that falls in the estimated time span $T_i$.
After filtering those documents, for each key event we get a new ranked list of documents ranked by their prediction scores from the document classifier.

\smallskip
\noindent \textbf{Iterative key event refinement.} We also introduce a feedback loop that iteratively refines the key event detection results using the current retrieved documents. 
First, given that phrase matching may generate some inaccurate pseudo-labeled documents, for each key event, we (1) remove its current pseudo-labeled documents that have negative prediction scores according to the trained classifiers and then (2) enrich the pseudo labels by adding the top-$n$ retrieved documents in the current ranked list of event documents (we set $n=5$ in our experiments). 
Second, we observe that the filtered document clusters in the \emph{post-classification processing} step may represent some possibly missing key events, because they are temporally close as they are clustered by publication times and semantically similar as they scored positive by the same classifier. 
Thus, we treat them as potential key events and construct their event-indicative features by applying standard tf-idf scores to select representative phrases from the cluster for pseudo label generation and using document publication times in the cluster as estimated event times. 
Finally, after refining the pseudo labels for current key events and also adding new event-indicative features, we can repeat the document retrieval process and iteratively improve the key event document selection results.

\section{Experiments}

\subsection{Experimental Setup}

\begin{table}[t!]
\centering
\caption{Datasets statistics.}\label{table:dateset}
\vspace*{-1.5em}
\scalebox{0.82}{
    \small
    \begin{tabular}{cccccc}
        \toprule
        \textbf{Dataset} & \textbf{\# Docs} & \textbf{\# Sents/Doc} & \textbf{\# Words/Doc} & \textbf{\# Events} & \textbf{\# Docs/Event}\\
        \midrule
        \textbf{HK Protest} & 1675 & 32.8 & 653.4 & 36 & 14.0 \\
        \midrule
        \textbf{Ebola} & 741 & 25.2 & 554.4 & 17 & 43.6 \\
        \bottomrule
    \end{tabular}
}
\vspace*{-2em}
\end{table}

\begin{table*}[!t]
\centering
\caption{Evaluation results on Ebola and HK Protest datasets using top-5 and top-10 documents for each predicted event. As our framework has randomness in the negative sampling process, we run it for 10 times and report the average of each measure.}
\label{table:main_res}
\vspace*{-1em}
\scalebox{0.9}{
\begin{tabular}{c|cccccc|cccccc}
\toprule
\multirow{2}{*}{Methods}    & \multicolumn{6}{c}{Ebola}                  & \multicolumn{6}{c}{HK Protest}                     \\
\cmidrule{2-7} \cmidrule{6-13}
                                                & 5-prec    & 5-recall    & 5-F1   & 10-prec & 10-recall & 10-F1 & 5-prec    & 5-recall    & 5-F1   & 10-prec & 10-recall & 10-F1 \\
\midrule
newsLens~\cite{Laban2017newsLensBA}             & 0.481 & \bf 0.765 & 0.591 & 0.524 & 0.647 & 0.579 & 0.352 & \bf 0.886 & 0.504 & 0.571 & 0.343 & 0.429 \\
Miranda et al.~\cite{Miranda2018MultilingualCO} & 0.444 & 0.706 & 0.545 & 0.733 & 0.647 & 0.688 & 0.481 & 0.371 & 0.419 & 0.286 & 0.057 & 0.095 \\
Staykovski et al.~\cite{Staykovski2019DenseVS}  & 0.414 & 0.706 & 0.522 & 0.688 & 0.647 & 0.667 & 0.442 & 0.657 & 0.529 & 0.444 & 0.114 & 0.182 \\
S-BERT                                          & 0.545 & 0.706 & 0.615 & 0.833 & 0.588 & 0.689  & 0.522 & 0.657 & 0.582 & 0.500 & 0.257 & 0.340 \\
\midrule
\OurNoClass                                     & 0.799 & 0.612 & 0.693 & 0.764 & 0.494 & 0.600 & 0.750 & 0.583 & 0.656 & 0.750 & 0.417 & 0.536 \\
\OurCOOC                                        & \bf 0.846 & 0.647 & 0.733 & \bf 0.909 & 0.588 & 0.714 & 0.815 & 0.611 & 0.698 & 0.807 & 0.431 & 0.561\\
\OurNoLM                                        & 0.784 & 0.659 & 0.715 & 0.865 & 0.635 & 0.732 & 0.905 & 0.608 & 0.728 & 0.942 & 0.453 & 0.612 \\
\OurSingle                                      & 0.814 & 0.671 & 0.735 & 0.872 & 0.635 & 0.734 & 0.916 & 0.636 & 0.751 & 0.958 & 0.458 & 0.620 \\
\Our                                            & 0.829 & 0.682 & \bf 0.748 & 0.883 & \bf 0.653 & \bf 0.751 & \bf 0.934 & 0.664 & \bf 0.776 & \bf 0.960 & \bf 0.464 & \bf 0.625\\
\bottomrule
\end{tabular}
}
\vspace*{-1em}
\end{table*}

\noindent \textbf{Datasets.}
We conduct our experiments on two real-world news corpora. Table~\ref{table:dateset} summarizes the statistics for these datasets. See Appx.~\ref{app:corpus_collection} for more details on theme corpus retrieval.
\begin{itemize}[leftmargin=*, nosep]
    \item \textbf{HK Protest}: We retrieve 1675 documents published from Mar. 21 to Nov. 21, 2019 about the theme ``2019 Hong Kong Protest'' from mainstream news publishers. There are 36 key events within our collected corpus. Note that more than half of the documents in this corpus do not belong to any of the 36 key events.
    \item \textbf{Ebola}: We obtain this dataset from the English part of the multilingual news clustering dataset published in~\cite{Miranda2018MultilingualCO}. Specifically, we select those events about the theme ``2014 Ebola Outbreak'' that have at least 20 supportive documents. The document publication times range from Sept. 18 to Oct. 31, 2014.
\end{itemize}

\smallskip
\noindent \textbf{Compared Methods.}
We compare the following methods for the key event detection task. See Appx.~\ref{app:imp_detail} for implementation details.
\begin{itemize}[leftmargin=*, nosep]
    \item {newsLens}~\cite{Laban2017newsLensBA}: A news stories visualization system that first clusters documents within several overlapped time windows and then merges them across time windows.
    \item {Miranda et al.}~\cite{Miranda2018MultilingualCO}: A multilingual online algorithm that either merges new documents with an existing cluster or creates a new cluster. It decides the document-cluster similarity using an SVM trained on labeled training documents.
    \item {Staykovski et al.}~\cite{Staykovski2019DenseVS}: A recent version of newsLens that explores different kinds of sparse and dense document representations.
    \item {S-BERT}: A variant of Staykovski et al. that uses the Sentence Transformer~\cite{reimers-2019-sentence-bert} to obtain document representations.
    \item {\OurNoClass}: An ablation of our framework that removes the document classification step and thus only uses the document-phrase matching results as the final output.
    \item {\OurCOOC}: An ablation of our framework using the raw document co-occurrence counts as edge weights between same-day peak phrases. The constant weight for same-phrase consecutive-day peak phrases is adjusted for the best performance.
    \item {\OurNoLM}: An ablation of our framework that removes the LM-based embedding similarity and only uses npmi scores as edge weights when constructing the peak phrase graph.
    \item {\OurSingle}: An ablation of our framework that only runs one iteration for the key event document selection module.
    \item {\Our}: Our full unsupervised key event detection framework.\footnote{\small Code and data can be found at:
    \url{https://github.com/yzhan238/EvMine}}
\end{itemize}

\smallskip
\noindent \textbf{Evaluation Metrics.}
Suppose there are $N$ ground truth key events $\G = \{G_1, G_2, \dots, G_N\}$ where each $G_i \in \G$ is a set of documents and $M$ predicted key events $\K = \{K_1, K_2, \dots, K_M\}$ where each $K_j \in \K$ is a ranked list of most confident event-related documents. Then, we think a predicted key event $K_j$ is \emph{K-Matched} to a ground truth event $G_i$, or $\text{KMatch}(K_j, G_i, k) = 1$, if more than half of $K_j$'s top-$k$ documents belong to $G_i$. Therefore, each key event can be \emph{K-Matched} to at most one ground truth event. Then, we define:
\begin{equation}
    \small
    \text{k-prec} = \frac{\sum_{G_i \in \G} \mathds{1}(\exists K_j, \text{KMatch}(K_j, G_i, k) = 1)}{\sum_{K_j \in \K} \mathds{1}(\abs{K_j} \ge k)},
\end{equation}
which is the number of distinct K-Matched ground truth events over the number of predicted key events with at least $k$ documents,
\begin{equation}
    \small
    \text{k-recall} = \frac{\sum_{G_i \in \G} \mathds{1}(\exists K_j, \text{KMatch}(K_j, G_i, k) = 1)}{N},
\end{equation}
which is the number of distinct K-Matched ground truth events over the number of ground truth events. Finally, k-F1 is the harmonic mean of the above two metrics.

\subsection{Experimental Results}

Table~\ref{table:main_res} shows that \Our can outperform all baseline methods on both Ebola and HK Protest corpora by a large margin in terms of 5-F1 and 10-F1 scores.
Such differences are more obvious on the HK Protest corpus, demonstrating that our method can deal with noisy corpus better than baselines.
Also, the S-BERT method brings some improvements compared to its base model, showing the effectiveness of our Sentence Transformer based document embeddings.
We notice some baseline methods can get overly high precision or recall, and the reason is that they either (1) generate large but too few events to get high precision but very low recall, or (2) generate small but too many events so that the recall is high but the precision will be too low due to duplicated events.

Our framework also outperforms all ablations. First, \Our can beat \OurNoClass which only finds documents within the estimated time period using document-phrase matching and skips those documents either outside the detected time periods or using different wordings than the detected peak phrases. In comparison, \Our trains text classifiers with automatically generated pseudo labels to retrieve more event-related documents for each key event. Second, \Our can outperform \OurCOOC which uses document co-occurrences when constructing the peak phrase graph. Although popular in previous related works~\cite{ge-etal-2016-event}, such a method can mix-up temporally overlapped events since they can be mentioned by the same document, while \Our uses npmi score that is better at distinguishing temporally similar events. Thrid, \Our can outperform \OurNoLM, especially on the recall metrics, showing that leveraging pretrained MLM-based embeddings helps identifying important synonyms that can be missed by the npmi scores. Finally, \Our can also outperform its single-pass version on all metrics, meaning that the iterative process can find some missing events and refine the document selection results.

\begin{figure}[t]
     \centering
     \begin{subfigure}{0.22\textwidth}
         \centering
         \includegraphics[width=\textwidth]{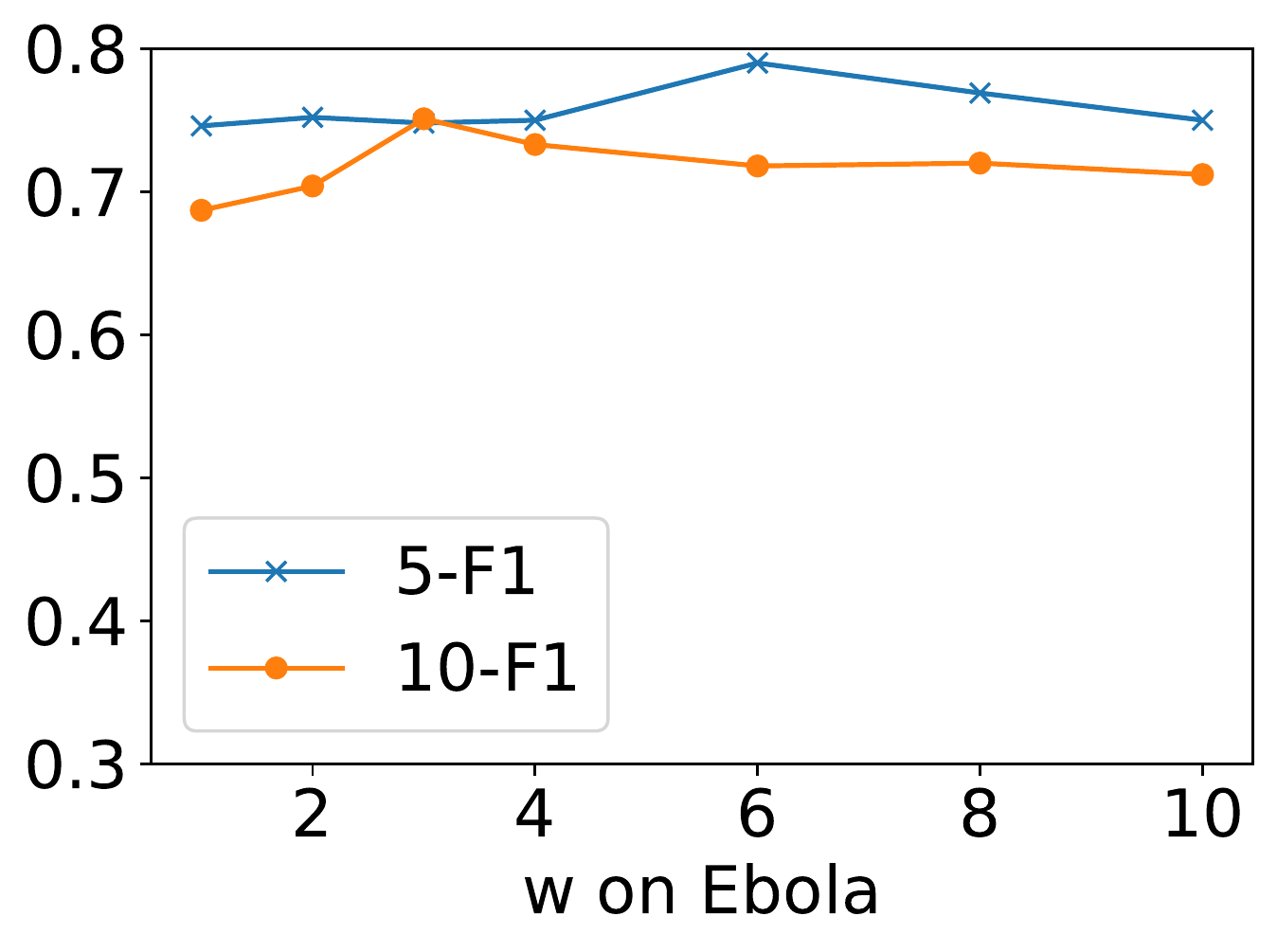}
        \vspace*{-1em}
         \label{fig:para1}
     \end{subfigure}
     \hfill
     \begin{subfigure}{0.22\textwidth}
         \centering
         \includegraphics[width=\textwidth]{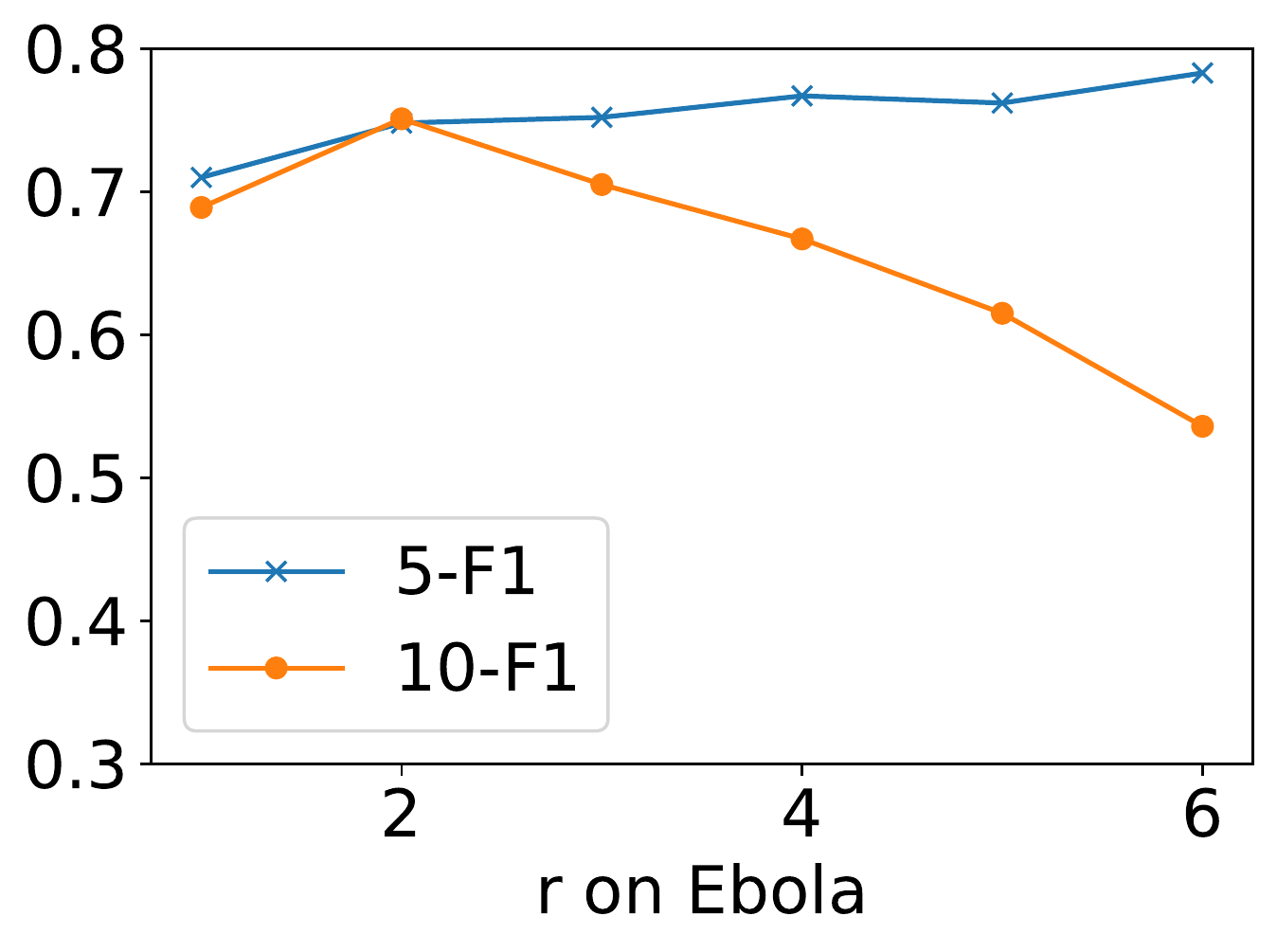}
        \vspace*{-1em}
         \label{fig:para2}
     \end{subfigure}
     \hfill
     \begin{subfigure}{0.22\textwidth}
         \centering
         \includegraphics[width=\textwidth]{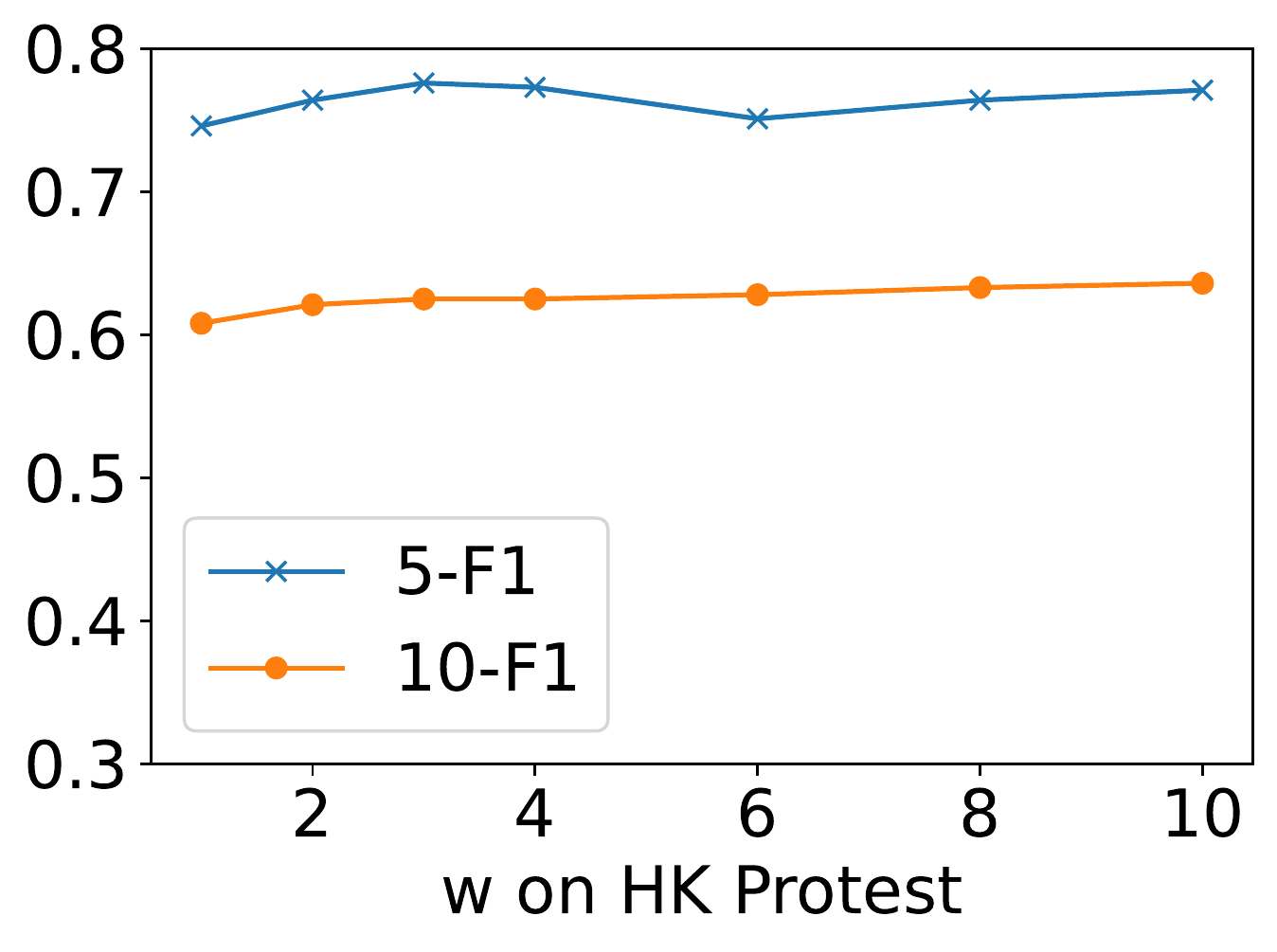}
        \vspace*{-1em}
         \label{fig:para3}
     \end{subfigure}
     \hfill
     \begin{subfigure}{0.22\textwidth}
         \centering
         \includegraphics[width=\textwidth]{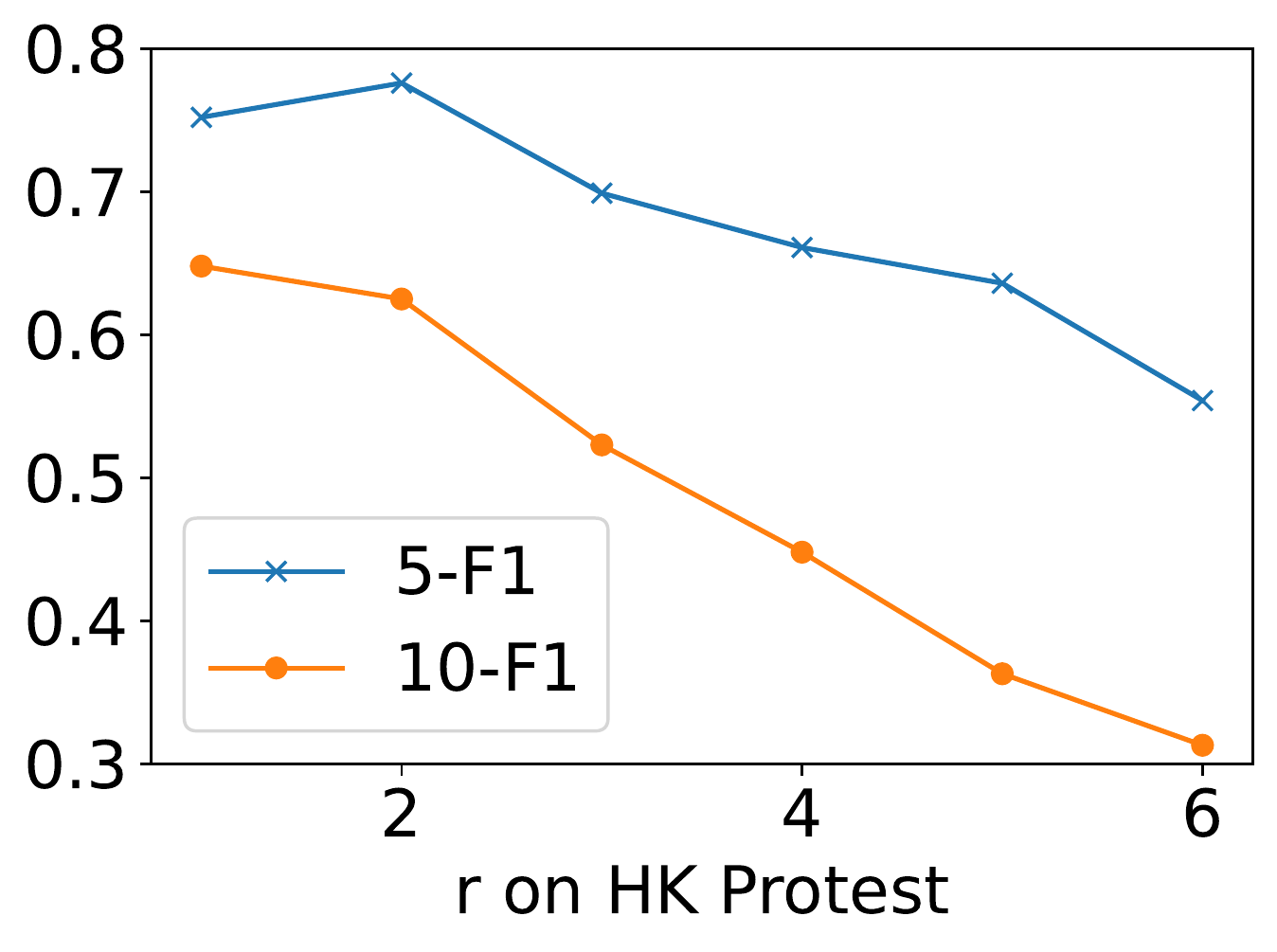}
        \vspace*{-1em}
         \label{fig:para4}
     \end{subfigure}
     \vspace*{-1.5em}
    \caption{Performances of \Our on the Ebola (top) and HK Protest (bottom) corpus when varying $w$ (left) or $r$ (right).}
    \label{fig:para}
     \vspace*{-1.5em}
\end{figure}

\subsection{Parameter Study} \label{sec:para_study}

In this section, we study how varying the following two parameters will affect the performance of \Our: (1) $w$, the constant edge weight between consecutive-day peak phrases with the same phrase for peak phrase graph construction (c.f. Sect.~\ref{sec:cluster}), and (2) $r$, the negative sampling ratio when training the text classifiers (c.f. Sect.~\ref{sec:doc_class}).
Figure~\ref{fig:para} shows the performance of \Our on Ebola and HK Protest corpora when varying these parameters individually. We can see that \Our is not very sensitive to the value of $w$. However, we still observe some performance drop on Ebola as $w$ becomes larger. The reason is that putting more weights on temporally closer peak phrases resulting in more thematically similar peak phrases to be separated into different communities, which makes the precision drop (same event being split) a little faster than the increase in recall (previous wrongly merged event being split). For the parameter $r$, the performance maintains stable for smaller $r$ and then starts to drop rapidly as $r$ reaches 3, especially for 10-F1. The reason is that larger $r$ means more negative samples used in classifier training and the model will overfit on the pseudo-labeled positive documents. This will result in fewer predicted documents for each key event and decrease the recall for the top-k evaluation metrics, since fewer events have more than $k$ documents. 

\subsection{Case Studies} \label{sec:case_study}

\begin{figure*}[!t]
    \centering
    \centerline{\includegraphics[width=0.95\textwidth]{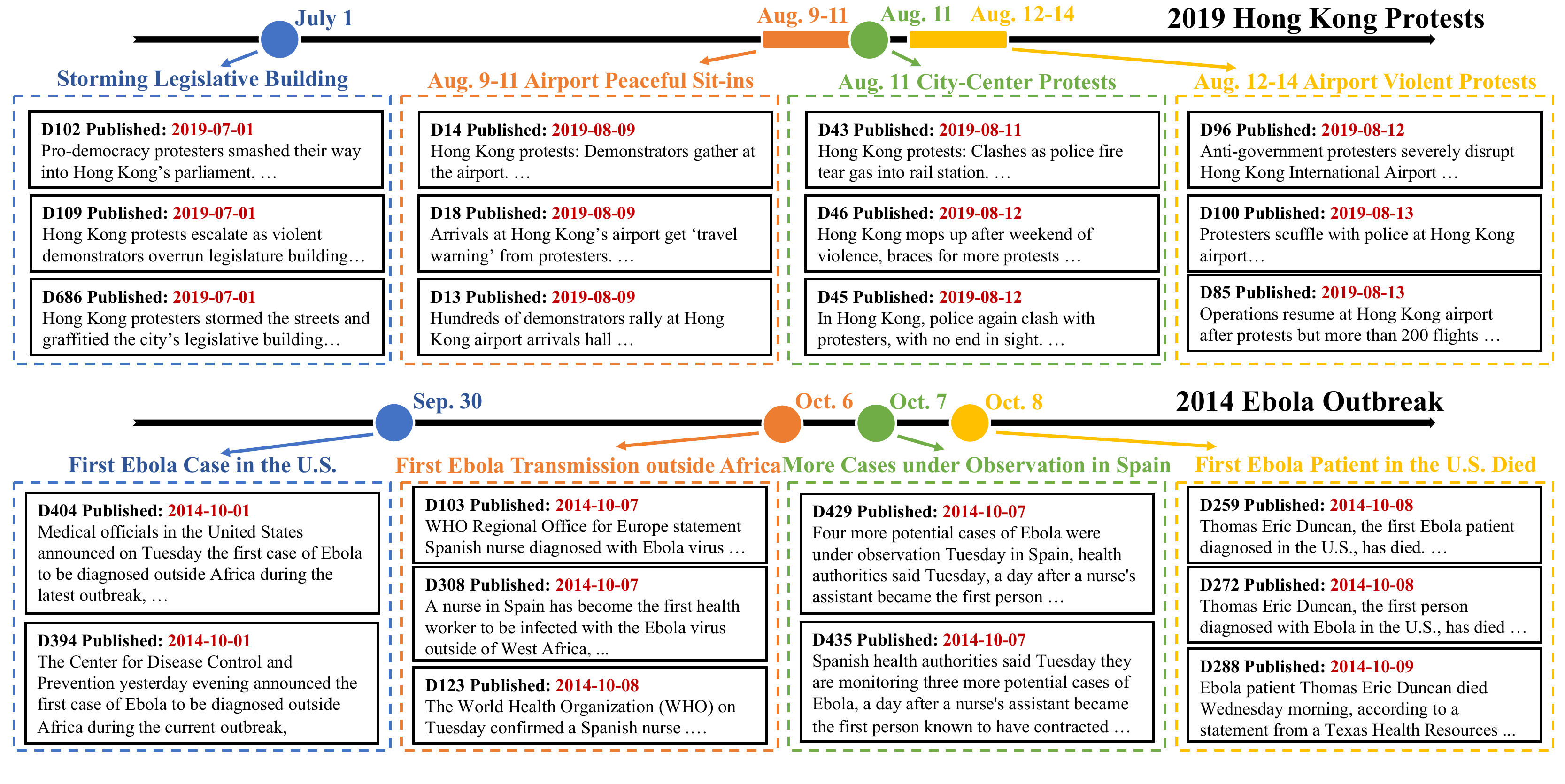}}
    \vspace*{-2em}
    \caption{Example outputs of our framework on HK Protest (top) and Ebola (bottom), including the titles (or the first sentences for Ebola) of some top-ranked event-related news articles predicted by our method that matched with real-world key events.}
    \label{fig:hk_cases}
    \vspace*{-1.5em}
\end{figure*}

\begin{figure}[!t]
    \centering
    \centerline{\includegraphics[width=0.45\textwidth]{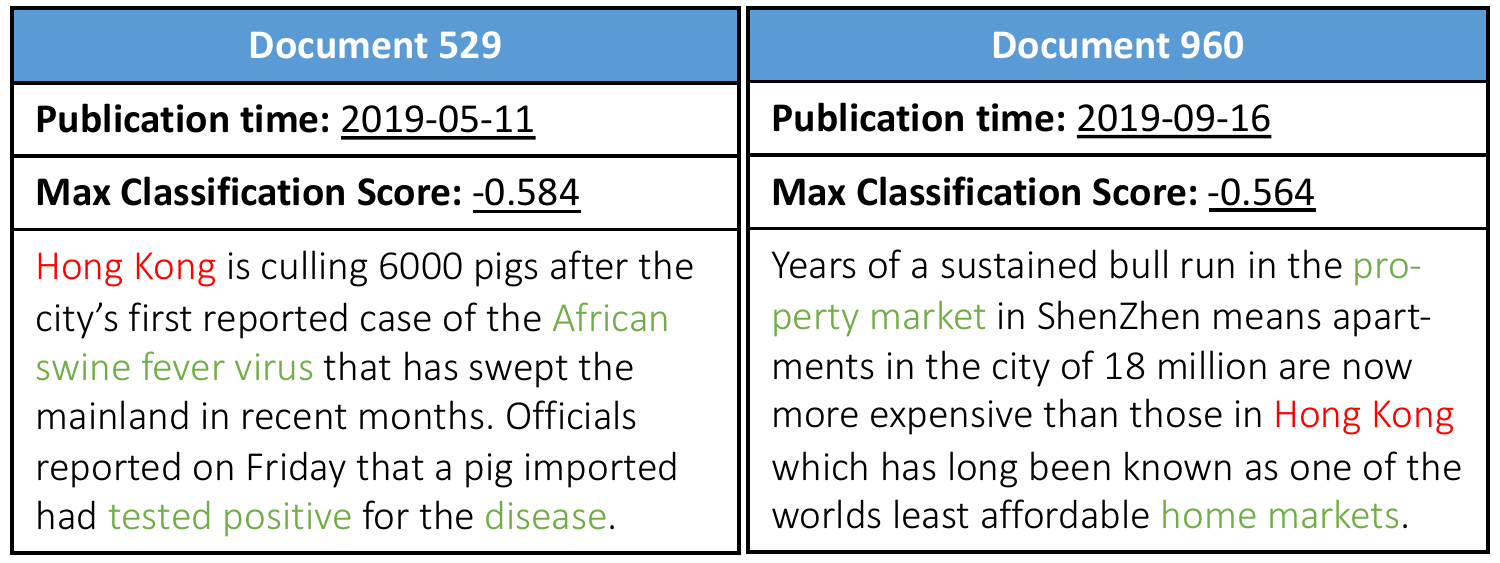}}
    \vspace*{-1em}
    \caption{Outlier documents found in HK Protest. Key phrases related to the major theme are colored \red{red} and those related to documents' actual topics are colored \green{green}.}
    \label{fig:outlier_cases}
    \vspace*{-1.5em}
\end{figure}

Figure~\ref{fig:hk_cases} shows several key events detected by \Our from HK Protest and Ebola.
In HK Protest corpus, we observe that \Our can distinguish temporally overlapped events (\textsc{Aug. 9-11 Airport Sit-ins} and \textsc{Aug. 11 Protests}) and thematically similar events (\textsc{Aug. 9-11 Airport Sit-ins} and \textsc{Aug. 12-14 Airport Protests}), demonstrating the effectiveness of our graph-based event detection algorithm that combines thematic and temporal information. Also, the high quality of the top-ranked documents for each detected key event shows the power of iterative document selection step. For Ebola corpus, \Our can distinguish temporally close key events (the three events happening in Oct.) and generate semantically coherent clusters of documents that align well with the ground truth key events.
Figure~\ref{fig:outlier_cases} additionally shows 2 documents from HK Protest corpus that have low document classification scores to all the detected key events. Although these 2 documents contain the key phrase ``Hong Kong'', their actual topics are not related to the major theme ``2019 Hong Kong Protests'', showing that EvMine can also identify outliers in the corpus and thus differs from simply clustering all the documents of a single theme.

\section{Related Work}

\noindent \textbf{Theme Detection.}
Topic Detection and Tracking (TDT), which aims to cluster news articles into thematic topics, is one of the earliest studies on automated event theme detection from text data~\cite{Allan1998TopicDA}.
One line of work~\cite{zhou2015unsupervised,Beykikhoshk2018DiscoveringTS} frames TDT as a topic modeling problem and addresses it using LDA~\cite{Blei2003LatentDA} or its improved variants.
Another line of studies~\cite{Sayyadi2009EventDA, Sayyadi2013AGA} builds a keyword graph using keyword co-occurrence statistics and discovers themes from clustered keyword communities. Then the topics are assigned to each document by their similarity with keyword-based feature vectors.
However, all these methods fail to consider the temporal information and thus can only distinguish thematically distinct events. 
To address this issue, some studies~\cite{Fung2005ParameterFB, He2007UsingBT} propose to model keyword burstiness for event detection.
Specifically, \cite{Zhao2012ANB} adopts the burst detection algorithm~\cite{Kleinberg2004BurstyAH} to extract bursty terms and uses them to represent and cluster documents.
Later, \cite{ge-etal-2016-event} constructs Burst Information Network using the detected bursts and their co-occurring features and proposes a node-based and an area-based clustering algorithms to detect events.
These studies have to first extract bursty terms, which can only be easily accomplished on corpus where events have huge theme-wise distinctions. 
In contrary, our key event detection task assumes the input news corpus is about one major theme, and the target key events may last as short as one day, making it hard to detect bursts for a single word or phrase.

\smallskip
\noindent \textbf{Action Extraction.}
Action extraction, also known as event mention extraction, is a popular NLP task aiming at extracting mention-level action consisting of a trigger and several arguments.
Traditional methods on this task~~\cite{Ji2008RefiningEE, gupta-ji-2009-predicting} rely on human selected linguistic features.
Later, neural models are also applied to automatically learn features for action extraction~\cite{Chen2015EventEV, Nguyen2016JointEE, Nguyen2018GraphCN}.
With the advancements on pre-trained language models (LMs), some studies also apply them on the action extraction task by either directly training LMs as trigger and argument extractors~\cite{Yang2019ExploringPL} or using LMs in a question answering way~\cite{Du2020EventEB}.
Compared to action extraction that aims to extract actions on the mention-level, our key event detection task focuses more on the super-event extraction at document-level by obtain key event indicative document clusters.
Also, while most event mention extraction methods are supervised and rely on massive labeled data, our proposed algorithm is unsupervised and can automatically detect key events without any human annotations.

\smallskip
\noindent \textbf{News Story Clustering.}
Some recent studies aim to detect stories from a general news corpus.
newsLens~\cite{Laban2017newsLensBA} is a news story visualization system that clusters documents within some overlapped time windows based on keywords and then links and merges the clusters across those windows.
Later, \cite{Staykovski2019DenseVS} improves newsLens by replacing keyword-based features with document representations.
\cite{Linger2020BatchCF} further extends~\cite{Staykovski2019DenseVS} to multilingual setting by training an S-BERT~\cite{reimers-2019-sentence-bert} model for document embeddings.
Besides the above batch clustering algorithms, \cite{Miranda2018MultilingualCO} proposes an online algorithm that compares new documents with existing clusters to merge them or create a new cluster.
The work also proposes a new multilingual news clustering dataset based on~\cite{Rupnik2017NewsAL}, from which we collect our Ebola corpus.
Recently, \cite{Saravanakumar2021EventDrivenNS} improves~\cite{Miranda2018MultilingualCO} by incorporating external entity knowledge into the contextualized representation of documents.
These studies, however, all assume each input document comes from exactly one real-world event, which may not hold for many real-world applications as the input corpus is often noisy.
In contrary, we propose \emph{key event detection} which aims to detect several most important events from a news corpus about one theme, and the corpus can be noisy in the sense that some documents may be related to multiple events or no event.

\section{Conclusion}

In this paper, we define a new key event detection task aiming at detecting real-world key (super) events given a news corpus about a general theme (\eg, detecting the event \textsc{July 1 Storming Legislative Building} for the theme ``2019 Hong Kong Protests'').
Compared with previous similar tasks that either study theme discovery or mention-level atomic event extraction, this new task is more useful for helping people gain insights to real-world events.
It is also inherently more challenging as the key events are thematically and temporally closer to each other and no labeled data is available for training given the fast evolving nature of news data.
To tackle this problem, we propose an unsupervised framework, \Our, that (1) defines a new metric ttf-itf to extract temporally frequent peak phrases from the corpus, (2) groups peak phrases by clustering on a novel graph structure that seamlessly combines topical, semantic, and temporal similarities, and (3) iteratively retrieves event related documents by training text classifiers with automatically generated pseudo labels and using the current results to refine the key event features.
Experiments on two real-world datasets show the effectiveness of our method on extracting key events even if they are thematically or temporally hard to distinguish.
In the future, we may explore the following directions: (1) currently the detected key events only have several documents and an estimated event time, so it would be interesting to automatically extract their actual times/locations and name the events; (2) by grouping documents into key events, we may investigate how to generalize event mention schema across different event types; (3) we can also further model the inner structure of key events as episodes (sub-events) to construct the entire event structure hierarchy.

\begin{acks}
Research was supported in part by US DARPA KAIROS Program No. FA8750-19-2-1004 and INCAS Program No. HR001121C0165, National Science Foundation IIS-19-56151, IIS-17-41317, and IIS 17-04532, and the Molecule Maker Lab Institute: An AI Research Institutes program supported by NSF under Award No. 2019897, and the Institute for Geospatial Understanding through an Integrative Discovery Environment (I-GUIDE) by NSF under Award No. 2118329. Any opinions, findings, and conclusions or recommendations expressed herein are those of the authors and do not necessarily represent the views, either expressed or implied, of DARPA or the U.S. Government.
\end{acks}

\bibliographystyle{ACM-Reference-Format}
\balance
\bibliography{ref}

\clearpage
\appendix
\section{NPMI Score Calculation}
\label{app:npmi}

The npmi between two same-day peak phrases $n_i = (p_i, t_i)$ and $n_j = (p_j, t_i)$, $t = t_i = t_j$, is defined as:
\begin{equation}
    \small
    \text{npmi}(n_i, n_j) = - \frac{\log \frac{\mathbb{P}(p_i, p_j, t)}{\mathbb{P}(p_i, t)\mathbb{P}(p_j, t)}}{\log \mathbb{P}(p_i, p_j, t)},
\end{equation}
where
\begin{equation}
    \small
    \mathbb{P}(p_i, p_j, t) = \frac{\abs{\{d \in \D | freq_d(p_i) > 0, freq_d(p_j) > 0, t(d) = t\}}}{\abs{\{d \in \D | t(d) = t\}}}
\end{equation}
\begin{equation}
    \small
    \mathbb{P}(p, t) = \frac{\abs{\{d \in \D | freq_d(p) > 0, t(d) = t\}}}{\abs{\{d \in \D | t(d) = t\}}}
\end{equation}
are the estimated probability of a document mentioning both phrases $p_i$ and $p_j$ on the day $t$, and the estimated probability of a document mentioning the phrase $p$ on the day $t$, respectively.

\section{Data Collection and Labeling}
\label{app:corpus_collection}

As discussed in Section~\ref{sec:retrieval}, for the HK Protest corpus, we use ``Hong Kong'' and ``protest'' as query to retrieve news articles from a massive news corpus crawled from mainstream news publishers and prune the articles if they are not published from Mar. 21 to Nov. 21, 2019. 
The result contains 1675 documents. 
To obtain key events within this corpus, three annotators are given the reference theme timeline ``2019 Hong Kong Protest''
(\url{https://en.wikipedia.org/wiki/Timeline_of_the_2019-2020_Hong_Kong_protests})
and asked to read each document to determine whether it describes a key event. 
Then the annotators need to discuss their decision and reach an agreement. 
We neglect those key events which are mentioned by too few documents. Finally, 36 qualified key events are discovered within our collected corpus.

To test the performance of our theme corpus retrieval step, we apply our theme corpus retrieval method on the news clustering dataset~\cite{Miranda2018MultilingualCO} to retrieve the documents about ``2014 Ebola Outbreak'' and compare the results with the documents selected according to their labeled events. 
Specifically, we use ``Ebola'', ``outbreak'' and ``medical'' as query while pruning the articles if they are not published in 2014. 
By comparing with the selected labeled documents about the theme ``2014 Ebola Outbreak'', we achieve recall = 0.982 and precision = 0.914, demonstrating our theme corpus retrieval method is simple yet effective.

\section{Experiment Implementation Details}
\label{app:imp_detail}

As the codes of newsLens~\cite{Laban2017newsLensBA} and Staykovski et al.~\cite{Staykovski2019DenseVS} are not published, we re-implement these two systems to the best of our knowledge.
For newsLens, we set the threshold for assigning edges $(T2)$ and the threshold for merging $(T3)$ to be 3 and 0.8, respectively.
For Staykovski et al., we choose $T2$ to be 0.7 and $T3$ to be 0.7. 
For S-BERT, we let $T2=0.7$ and $T3=0.8$.
For all the above three systems, the time interval is 3 and the window overlap is 2; then within each cluster, documents are ranked by their number of edges in the graphs.
For Miranda et al.~\cite{Miranda2018MultilingualCO}, since the authors do not publish the SVM training code, we test their published pre-trained model
on both corpora
(\url{https://github.com/Priberam/news-clustering}).
For \Our, the constant edge weight $w$ between same-phrase consecutive-day peak phrases is set to 3, the negative sampling ratio $r$ for classifier training is 2, and the number of classifiers $S$ trained for each key event is 50.
We provide a parameter study on $w$ and $r$ in Section~\ref{sec:para_study}. Besides, when calculating the ttf score, we set the number of afterward days $n_t$ for frequency aggregation to 3, and our method is not sensitive to this parameter as long as it is greater than 2. We obtain the MLM-based phrase embeddings using BERT-base-uncased model, and the threshold $\tau$ for event-indicative phrases enrichment (c.f. Sect.~\ref{sec:doc_class}) is set to 0.95.
We extract time expressions from the news articles using the datefinder package
(\url{https://github.com/akoumjian/datefinder}).
For the iterative refinement, every time we add the retrieved top-5 (\ie, $n=5$) documents to the pseudo labels for each key event. We run 2 iterations in our experiments and empirically find that 2 to 3 iterations are enough to yield the best performance.
\end{document}